\documentclass{article}

\usepackage[preprint,nonatbib]{nips_2018}

\usepackage[utf8]{inputenc} 
\usepackage[T1]{fontenc}    
\usepackage{hyperref}       
\usepackage{url}            
\usepackage{booktabs}       
\usepackage{amsfonts}       
\usepackage{nicefrac}       
\usepackage{microtype}      
\usepackage{listings}
\usepackage[utf8]{inputenc}
\usepackage{graphicx}
\usepackage{wrapfig}
\usepackage{amsmath,amsfonts,amssymb,amsthm}
\usepackage{subcaption}

\usepackage{color}
\usepackage{soul}  

\graphicspath{{images/}}

\begin{document}

\title{Distilling Spikes: Knowledge Distillation in \\ 
Spiking Neural Networks}
\author{
  Ravi Kumar Kushawaha{$^*$} ,
  Saurabh Kumar{$^*$} ,
  Biplab Banerjee,
  Rajbabu Velmurugan
  \\[8pt]
Indian Institute of Technology Bombay \\
\texttt{rkkush2397@gmail.com, \{saurabhkm, bbanerjee, rajbabu\}@iitb.ac.in}\\
}

\maketitle
\begin{abstract}
\renewcommand{\thefootnote}{*}
\footnotetext{Equal Contributions}
Spiking Neural Networks (SNN) are energy-efficient computing architectures that exchange spikes for processing information, unlike classical Artificial Neural Networks (ANN). Due to this, SNNs are better suited for real-life deployments. However, similar to ANNs, SNNs also benefit from deeper architectures to obtain improved performance. Furthermore, like the deep ANNs, the memory, compute and power requirements of SNNs also increase with model size, and model compression becomes a necessity. Knowledge distillation is a model compression technique that enables transferring the learning of a large machine learning model to a smaller model with minimal loss in performance. In this paper, we propose techniques for knowledge distillation in spiking neural networks for the task of image classification. We present ways to distill spikes from a larger SNN, also called the teacher network, to a smaller one, also called the student network, while minimally impacting the classification accuracy. We demonstrate the effectiveness of the proposed method with detailed experiments on three standard datasets while proposing novel distillation methodologies and loss functions. We also present a multi-stage knowledge distillation technique for SNNs using an intermediate network to obtain higher performance from the student network. Our approach is expected to open up new avenues for deploying high performing large SNN models on resource-constrained hardware platforms.
\end{abstract}

\section{Introduction}
Neural Networks have become immensely popular due to the increase in their performance, research interest, accessibility of computing infrastructure and development tools. They currently are the state-of-art algorithms for problems from diverse fields ranging from pattern recognition in computer vision and language processing to applications in computational sciences. However, most of these architectures are made out of the classical artificial perceptron, which works in an analog fashion. These architectures work as non-linear continuous function approximators operating synchronously with a common clock cycle. This makes them power-hungry and tedious for real-world deployment.

Neural networks, in general, are known to be inspired by the biological brain in a manner that each perceptron behaves similar to a biological neuron and is connected to various other perceptrons forming a large network. Each of these individual perceptrons, take inputs from various other perceptrons and produce an output. However, the perceptrons in an ANN work in an analog manner, unlike a neuron in the brain. A biological neuron in the brain only fires for a very short duration when provided with appropriate stimuli and then stops. These sparse and asynchronous pulses for information exchange means the neurons mostly remain inactive and hence require significantly lower energy \cite{paugam2012computing}. These pulses are also called spikes, and spiking neural networks attempt to model this behavior for developing more efficient computing architectures. These architectures can be deployed on neuromorphic hardware and have been shown to demonstrate extremely low energy consumption in contrast to the ANNs \cite{diamond2016comparing}. SNNs on neuromorphic hardware provide very fast inference and event-driven processing with low power requirements.

\begin{figure}
    \centering
    \includegraphics[width=\columnwidth]{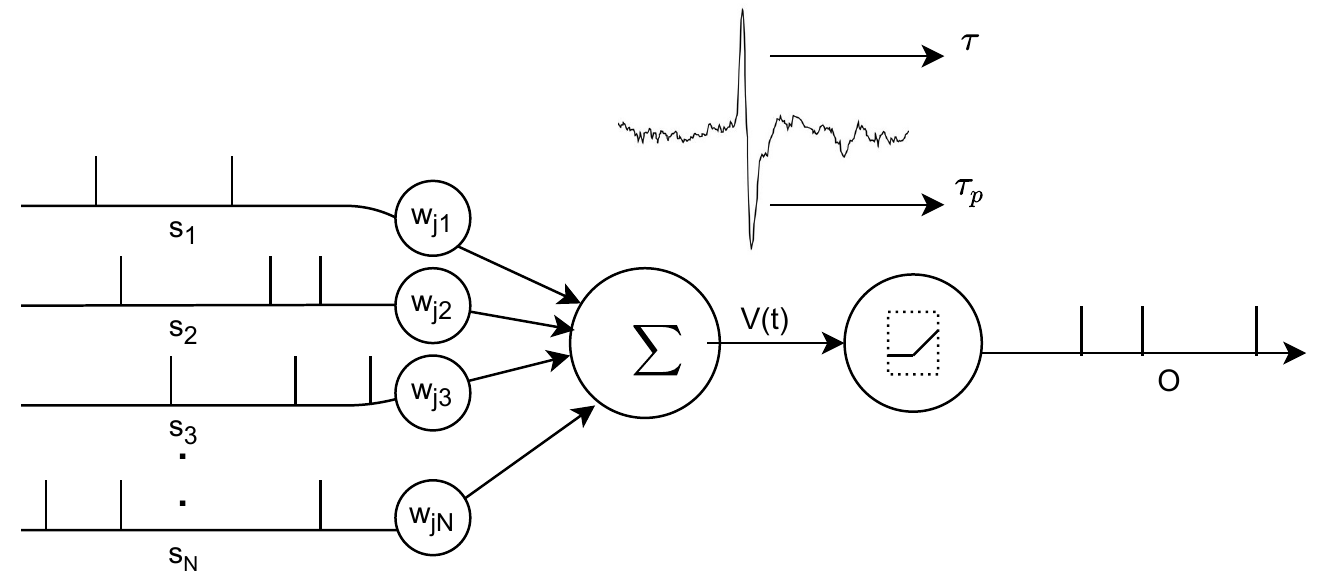}
    \caption{An illustration of the working of a spiking neuron. Input spike trains $s_i(t)$ are weighted by synaptic weights $w_i$ and summed to obtain the membrane potential $V(t)$. The neuron generates a spike if $V(t)$ goes above a threshold $\tau$ and resets to baseline after a refractory period modelled using a penalty threshold $\tau_p$.}
    \vspace{-6mm}
    \label{fig:spikingNeuron}
\end{figure}

Early works include work by Gerstner et al. \cite{gerstner2002spiking} who studied SNNs as composed of neurons that exchange information via spikes as biological information processing models. Information is exchanged via Spikes that are asynchronous events and therefore have a time dimension for the spiking frequency associated with them, unlike the logits in ANNs. An illustration of the working of a single spiking neuron is shown in Figure \ref{fig:spikingNeuron} for clarity.

The input spikes to a neuron are integrated as membrane potential, and once it reaches a threshold $\tau$, the neuron generates an output spike. We can describe this model as follows. Consider a spike train $s_i(t)$ which is multiplied by synaptic weights $w_i$'s to produce post-synaptic potential for each of the $i^{th}$ input:
\begin{equation}
    V(t) = \sum_{i=1}^{N} w_i s_i(t)
\end{equation}
where $V(t)$ is called the membrane potential, and $w_i$'s are the weights of the neuron. The neuron fires when the sum of post-synaptic potentials i.e. $w_i * s_i(t)$ goes above a threshold $\tau$, i.e., an output spike is generated when $V(t) > \tau$. The values accumulated during this pre-activation phase exponentially decays at every time step until the neuron fires. This spike is sent to all connected neurons, and the membrane potential is reset to a baseline. This process is then repeated at every neuron. In addition to this, to prevent spurious spikes, there is a penalty threshold to reflect the refractory period where the neuron firing is inhibited once it goes below a certain threshold say, $\tau_p$.

However, large machine learning models pose significant problems in practical deployment in terms of memory, time, and computational power. Despite being powerful models offering state of the art performance, neural networks are huge models for deployment on edge devices, like sensor networks or smartphones. Real-world devices are limited in resources, and the model architectures need to be adapted for deployment on such devices. Reducing the model size is an essential practical problem in such scenarios, and Knowledge Distillation (KD) is a useful technique for this. KD is a technique that enables us to transfer the knowledge from a large network that has been trained to solve a certain task, to a smaller network \cite{hinton2015distilling}. As per the KD paradigm, knowledge in a network is not in the weights of the network. It is instead in the output activations produced when given a specific output. So if a smaller network can produce the same activations for a given input as the large trained network, then they both are solving the task equally well. In reality, this is not exactly doable due to a decrease in model capacity as the number of parameters reduces. However, for practical purposes, we might be willing to take a small hit in the performance while significantly reducing the model size.

KD has been extensively researched and effectively used for ANNs in the literature. However, some works show that KD is not always effective, especially when the size difference between the teacher model and the student model is large. Distilling knowledge from a large student model to a much smaller student model yields sub-par performance. This impacts the efficacy of the process as such different models do not form a perfect knowledge transfer pair. To mitigate this problem, an intermediate Teacher-Assistant network has been explored by Mirzadeh et al. \cite{mirzadeh2019improved}. Using a multi-step distillation process, they show that the student network learns better via an intermediate TA model than if it was trained directly from the teacher.

In this paper, we show how KD can be applied for training SNNs. We train the SNNs with back-propagation in a supervised manner. We show how off-the-shelf objective functions commonly used in ANNs can be adapted for SNNs. We demonstrate the effectiveness of the proposed methods with exhaustive experiments. With advances in neuromorphic hardware research and practical deployment of SNNs, we believe this work can be a step towards obtaining the excellent performance of deep SNNs while working in the physical constraints of the available hardware. To the best of our knowledge, this is the first work in the direction of distilling knowledge in SNNs models.

\subsection{Our Contributions}
\begin{itemize}
\item{We present the first-ever method to distill knowledge from a large SNN model trained for image classification to a smaller SNN model with minimal loss in performance.}
\item{We propose a novel training strategy and multiple objective functions to distill knowledge in SNNs.}
\item{We present a multistage knowledge distillation procedure suited for SNNs using an intermediate Teacher Assistant network to improve the performance of a smaller student network.}
\item{We demonstrate the effectiveness of the proposed spike distillation approach by thorough experiments on multiple datasets.}
\end{itemize}

\section{Related Work}
Considering the focus of the paper, here we discuss briefly past works on Spiking Neural Networks and Knowledge Distillation.

\subsection{Spiking Neural Networks}
SNNs have been a topic of active research with advances in architecture designs \cite{tavanaei2019deep}, training methodologies \cite{neil2016learning}, \cite{wu2019direct} and practical deployments \cite{ankit2017resparc}, \cite{lin2018mapping}. SNN architectures have started becoming more complex and larger to extract more performance. They have been successfully applied to diverse pattern recognition tasks from object recognition \cite{cao2015spiking} to EEG classification \cite{antelis2020spiking}. Recently there have been works that show SNNs achieve comparable performances as the deep neural network architectures like VGG16 and Res-Nets for image classification tasks \cite{sengupta2019going}. On similar lines, Kim et al. \cite{kim2019spiking} presented a first spike-based object detection model, called Spiking-YOLO.

Training SNNs is difficult as the spike events are not differentiable. There have been many techniques proposed to train SNNs. Two of the popular ones are by using back-propagation and Hebbian learning. Lee et al. \cite{lee2016training} proposed a modified back-propagation based approach where the membrane potentials are treated as differentiable signals, and discontinuities at spike times are considered as noise. There is also a training methodology that considers the relative timing between the pre-synaptic and post-synaptic spikes to influence the training of synaptic weights via an asymmetric learning window \cite{kempter1999hebbian}. Yin et al. \cite{yin2017algorithm} proposed an SNN training algorithm with continuous integration, which can handle input spikes with temporal information. There are also Hebbian learning \cite{gupta2009hebbian} based approaches that enable local training, unlike back-propagation.


\subsection{Knowledge Distillation}
Knowledge Distillation refers to the process of transferring knowledge of an already trained, large/complex learning model or an ensemble of several models to a rather small counterpart, which is expected to be easier to comprehend and deploy. Bucilua et al. first introduced knowledge distillation for machine learning models. \cite{bucilua2006model} and was later applied and popularized for deep neural networks by Hinton et al. \cite{hinton2015distilling}. Bucilua et al. \cite{bucilua2006model} use the final softmax outputs of a larger ANN to learn a smaller model by minimizing the squared difference between their activations. The method in \cite{hinton2015distilling} is a more general one where the output logits are scaled by a temperature parameter to obtain soft thresholds and then used in the loss. KD has been an active area of research with major applications in model compression \cite{cheng2017survey}. There have also been works that demonstrate improvement in the performance of a model when using KD based learning, e.g., in object detection \cite{chen2017learning}. In addition to this, KD has also been used to improve the performance of various pattern recognition tasks when used as a training methodology for language understanding in \cite{liu2019improving} and sequence models in \cite{huang2018knowledge}.

On similar lines, the work by Kim et al. \cite{kim2016sequence} proposed that instead of minimizing cross-entropy with observed data, it is better to minimize cross-entropy with teacher's probability distribution. Since it provides more information about other classes for a given data point and has less variance in gradients. A recent work by Mirzadeh et al. \cite{mirzadeh2019improved} presents improvements in knowledge distillation by using an intermediate network they call teacher assistant. Inspired from this work, we also present a novel KD procedure suited for SNNs and also develop a teacher assistant based multi-stage distillation procedure for SNNs in this paper. As described above, KD is an active and well-researched area for ANNs, but there are no such methods available for SNNs to the best of our knowledge. With the advent of larger SNNs models and their practical deployment on neuromorphic hardware, KD for SNNs is an essential and crucial prospective research direction to extract more performance while working with hardware constraints.

\section{Proposed Method}
\label{sec:proposedMethod}

\subsection{SNN model}
The architecture of an SNN is similar to traditional multi-layer fully connected feed-forward ANNs, even though the neurons work differently. Each spiking neuron receives a set of pre-synaptic spikes as input that are weighed and passed through a nonlinearity to produce post-synaptic spikes. Due to their design, SNNs have a time axis for representing inputs and outputs. This concept of time comes in as the spike signals are not a single scalar values, but a stream of spikes also referred to as a spike train. Therefore the representation of a relationship between the input and output values is different in SNNs when compared to ANNs.

As a first step, we have converted the input images into spike trains that the SNN can work with. For this, each two-dimensional input image is flattened to a one-dimensional vector, and the elements of the vector are fed into the input layer as constant spike trains. The post-synaptic spikes generated by the input layer are transmitted to the intermediate layers, whose input neurons and the output neurons are densely connected. The intermediate layers modulate the received spike trains by their weights to generate output spike trains, which in turn act as the pre-synaptic spikes for the next layers. Finally, the post-synaptic spikes of the intermediate layers are available to the output layer. The output layer similarly uses the spike train from the penultimate intermediate layer to generate the final output spike train that is used to infer the class of the input image sample. The neurons in each layer are connected via weights to the next higher layer in a fully connected manner. Note that within each layer, we have an extra dimension of time to represent the spike trains.

Each of the neurons receive inputs from the previous layer in the form of spike trains. These inner excitations are accumulated into a membrane potential until a threshold, $\tau$, is reached. A neuron generates a spike only after reaching this threshold. This input pulse contributes to the inner activation value to rise and then gradually declines. To avoid accumulation of inner firing stimuli for too long, the previous inner activations are decayed over time with a decay factor $\lambda$. The decay acts as an exponential moving average on the most recent values seen, and the inner excitations are then added to the value of previous inner activations decayed over several time steps. In effect, the inner excitation accumulates through time and also exponentially fades away. This way, during back-propagation, a stimulus which happened much longer back in time will suffer vanishing gradients and not contribute to the learning process as it is very likely not important. This can be written as,
\begin{equation}
    x = i + \lambda *x' 
\end{equation}
where $x$ is the inner excitation, $i$ is the input to the neuron, $x'$ is the value of inner excitation in the previous time step, and $\lambda$ is the decay multiplier.

If there is a spike, the inner excitation is reduced to a significantly low value so that the neuron does not fire until some time, i.e., refractory period. Outer excitation occurs only when a threshold is reached, this can be written as,
 \begin{equation}
     y = \sigma(x-\tau)      
 \end{equation}
where, $y$ is the outer activation, $x$ is the inner activation, $\tau$ is activation threshold and $\sigma$ is the activation function.
 
Whenever there is a spike, the inner excitation is penalized so as to act as a reset to prevent spiking again spuriously.
\begin{equation}
  p =
    \begin{cases}
      1 & \text{if $y$ $>$ 0}\\
      0 & \text{otherwise}
    \end{cases}       
\end{equation}

\begin{equation}
    x = x - p * (\frac{\tau_p}{\tau} * x)
\end{equation}
where $\tau_p$ is the self penalty threshold, $p$ is the penalize gate, which is a boolean depending on the outer activation.

From the architecture point of view, we additionally propose to use ReLU as the activation function and adding batch norm after each synapse.


 

\begin{figure*}
\begin{center}
   \includegraphics[width=\linewidth]{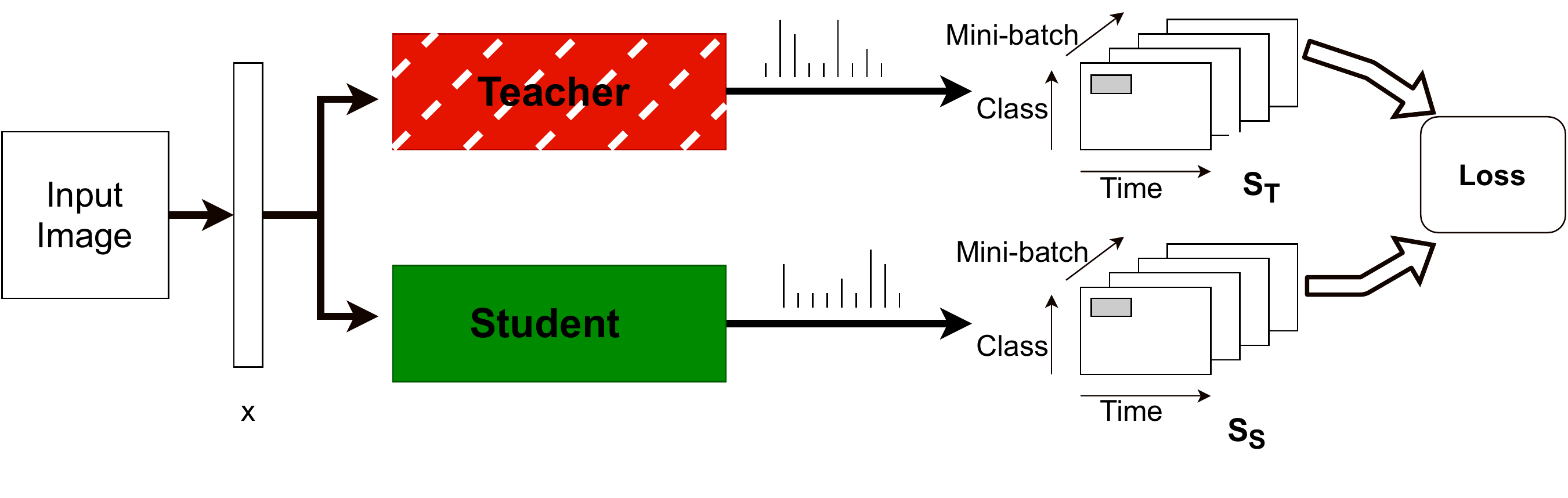}
\end{center}
   \caption{This figure shows the complete training methodology. We first train a teacher SNN which is then used in KD for a student network. The second step is the KD process, where, given an input image, the weights of teacher SNN are frozen (shown in the figure by dashed network block) while only the student SNN is trained. The KD process involves training this two-stream setup with the proposed loss functions on the post-synaptic spike patterns of the Teacher and Student SNNs.}
\label{fig:distillationProcess}
\end{figure*}

\subsection{SNN Network Architectures and Training}
In our experiments, the Teacher SNN is a large model, and the Student model is a much smaller SNN model. The input images are converted to spikes, and after that, log-softmax values are taken for each class to calculate the loss with ground truth labels. For training, the vectorized input image data, $I$, is fed to the SNN models. The output layer contains ten neurons corresponding to each of the ten classes in the datasets. The network is trained using back-propagation where the outputs are converted to softmax activations, which are then used to calculate Negative Log-Likelihood loss with respect to the target labels, as shown below
\begin{equation}
    Loss_{NLL} = - \sum_{B} y_{true}^{k} \hspace{5pt} log(\hat{y}_{true}^{k})
\end{equation}
where $y_{true}^{k}$ is the ground truth label and $\hat{y}_{true}^{k}$ is the softmax value of target label class.

\subsection{Proposed Knowledge Distillation Procedure}
Since the student relies on final layer spike train of the teacher with respect to the time axis, and not on the target labels with which the teacher was trained, the forward function is different for the teacher and student models for training. We use the output spike train before softmax and loss calculation for the KD process. During this, the weights of the teacher model are frozen, and its output spike trains for a given input sample are used to train the student model while providing it with the same input sample as the teacher model.

To do this, we also propose a novel loss function that enables the successful distillation of knowledge from a large SNN model to a smaller SNN model. Over each time step and mini-batch of the input, the final output post-synaptic spikes for each of the output neurons are arranged to construct a three-dimensional tensor that is used in our proposed distillation process. More precisely, both of the teacher and student models output a $t \times c \times b$ matrix each, where $t$ is the discrete number of steps in the time axis for each of the $c$ classes and a mini-batch size of $b$. We shall refer to this quantity as the spiking activation tensor and denote it by $\mathcal{S}$. Now, these spiking activation tensors are used to train the student model parameters from the outputs of the teacher model with the proposed loss functions described below.

At every step of training the student via the proposed KD approach, the spiking activation tensor (SAT) of a student model, denoted by $\mathcal{S}_{S}$ is compared with the spiking activation tensor of the frozen teacher model, denoted by $\mathcal{S}_{T}$. We propose the use of two sets of objective functions and find that a combination of both works the best in our experiments. For the first set of objectives, The full losses are calculated by comparing the full SATs of both teacher and student model. We use full L1, full L2 and KL losses which can be written mathematically as shown in Equations \ref{eq:fullLossCalulationKL} and \ref{eq:fullLossCalulation}. Along with this, we propose a second set of objectives which we call the sliding window losses that work in moving window fashion over the corresponding regions of SATs of teacher and student models. Given an input image sample, the teacher model generates an SAT we denote by $\mathcal{S}_{T}$, and the student model generates an SAT we denote by $\mathcal{S}_{T}$. In the sliding losses, for each class, over the T length vector, a sliding window of length $\Delta$ is taken, and loss is computed with respect to the corresponding sliding window in $\mathcal{S}_{S}$. The window is shifted, and this loss is accumulated over the total length of the vector. This can be mathematically written, as shown in Equation \ref{eq:slidingLossCalulation}. This computed loss is then back-propagated to train the parameters of the student network while keeping the teacher network frozen.

\begin{equation}
\begin{split}
L_{KL} & =  D_{KL}( \sigma_{softmax}(\mathcal{S}_{S}) || \sigma_{softmax}(\mathcal{S}_{T})) \\
 & = (\sigma_{softmax}(\mathcal{S}_{S}))*log\Big(\frac{\sigma_{softmax}(\mathcal{S}_{S})}{\sigma_{softmax}(\mathcal{S}_{T})}\Big)
 \label{eq:fullLossCalulationKL}
\end{split}
\end{equation}

\begin{equation}
L_{Lm} = ||\mathcal{S}_{T} - \mathcal{S}_{S}||_{m}
\label{eq:fullLossCalulation}
\end{equation}

\begin{equation}
L_{sLm} = \sum_{k \epsilon b} \sum_{j \epsilon c} \sum_{i \epsilon t} ||\mathcal{S}_{T}[i:i+\Delta;j;k] - \mathcal{S}_{S}[i:i+\Delta;j;k]||_{m}
\label{eq:slidingLossCalulation}
\end{equation}

\begin{equation}
Loss =  \alpha L_{sL1} + \beta L_{L2} + \gamma L_{KL}
\vspace{-10pt}
\end{equation}
\begin{eqnarray*}
\alpha , \beta , \gamma > 0 \\
\alpha + \beta + \gamma = 1
\end{eqnarray*}
where, $m$ represents the $m$-th norm, KLD represents the KL Divergence loss.
$\sigma_{softmax}$ is the softmax activation function. $\alpha$, $\beta$, and $\gamma$ represent the weights given to the losses L1, L2, and KL, respectively.

The proposed loss functions are designed such that the spike pattern of the student will closely follow the spike pattern of the trained teacher for a given sample, hence despite being a shallow network, it will try to mimic the spike train of a deeper network as best as possible. It is beneficial to guide the student SNN model towards a weight configuration that results in producing similar activations as the teacher SNN model to obtain effective distillation. We observe that the sliding window based $L1$ loss function achieved better performance than full $L1$ loss between the SATs. For $L2$ and $KD$, we observe that the full objectives provide better performance. While testing, we convert the spike trains of the student model to softmax activations and compare it with ground truth labels using NLL loss. 

\subsection{Training with Teacher Assistant Model}
Building on the work in \cite{mirzadeh2019improved}, we introduce a multi-stage distillation procedure to improve the performance of KD in SNNs. Mirzadeh et al. observed that if the gap between the size of the teacher model and the student model is large, the knowledge distillation process is not effective. As per their experiments, a student model trained from a much larger teacher model performs worse than a student model trained from a smaller teacher. To mitigate this problem, they propose an intermediate network that is of a size that is in between that of the teacher and student model to avoid the significant size difference. On similar lines, we propose an intermediate SNN model that acts as a teacher assistant to enable better distillation from more a massive teacher model to smaller student. In this we, instead of directly training student, the teacher first trains an intermediate model which we refer to as Teacher Assistant using the proposed KD loss. This TA model is subsequently used to train the student model using the same KD loss.

\section{Experimentation and Results}
In this section, we present the details of our experimental methodology using the proposed approach, along with datasets and results.

\subsection{Dataset Description}
We experiment with three standard image classification datasets for evaluating the effectiveness of our proposed knowledge distillation methodology. The datasets are MNIST, Fashion-MNIST, and CIFAR-10, and all of them consist of 10 classes. MNIST is a popular dataset used for benchmarking image classification tasks. It consists of 70000 images of handwritten digits of size $28 \times 28$. The dataset provides a train-test split consisting of 60000 samples for training and 10000 for testing. The second dataset we use is the Fashion-MNIST \cite{xiao2017fashion}, which is a dataset of Zalando's article images consisting of a training set of 60000 examples and a test set of 10000 examples. Here again, each example is a $28 \times 28$ grayscale image, associated with a label from 10 classes. The third dataset we use is the CIFAR-10 image dataset consisting of 60000 $32 \times 32$ color images of 10 natural objects. There are 50000 images for training and 10000 for testing, with 6000 images per class. The test set contains exactly 1000 randomly-selected images from each class. Between them, the training set contains exactly 5000 images from each class. The images were first converted to grayscale before use. We used the standard train-test splits provided for each of these datasets in our experiments.

\subsection{Experimental Protocol}
The network implementation uses Adam optimizer for training with a learning rate of $10^{-2}$. The threshold and penalty thresholds are kept at $1.0$ and $1.5$ respectively. Each of the models is trained for 100 epochs, and performance is measured using the overall accuracy of classification. The teacher model used in our experiments is of the largest size and consists of six SNN layers. The student model is the smallest and consists of two SNN layers. And finally, the intermediate TA model is chosen to be of four SNN layers. The sliding window size is varied from $32$ to $128$. And, the weights of the loss function components are obtained using Grid search.

\subsection{Experimental Results}
We evaluate the performance of the proposed KD techniques and objective functions via multiple experiments on the above three datasets, and the results are presented in this subsection.

We train the Teacher network using the proposed method described in section \ref{sec:proposedMethod}. We observe that MNIST is most easy to learn, whereas Fashion-MNIST and CIFAR10 are more difficult to obtain good performance on. These datasets are used extensively in the literature for robust evaluation of learning algorithms. We obtain classification accuracies comparable to the fully connected ANN architectures for each of the datasets. Moreover, the classification performance with our training and distillation procedure reflects the same trend.

For the first experiment, we establish the baseline performances to demonstrate the effectiveness of our proposed method. We individually train the Teacher(T), Teacher Assistant(TA), and Student(S) SNN models from scratch on each of these three datasets. The results of this experiment are presented in table \ref{tab:individualPerformance}. We can observe from the table that the TA model due to its lower model capacity achieves a performance lower than the larger teacher network. The student model achieves an even lower performance due to an even smaller number of trainable parameters in it. We shall use these as our benchmark to compare the upcoming experiments.

\begin{table}[!t]
\caption{Baseline classification performances of individual networks when trained separately on the three datasets.}
    \label{tab:individualPerformance}
    \centering
    \begin{tabular}{c|c c c}
        \hline \textbf{Dataset} & \textbf{MNIST} & \textbf{F-MNIST} & \textbf{CIFAR10}\\ \hline
        Teacher & 98.35 & 89.72 & 45.43 \\
        TA & 98.17 & 89.4 & 45.98 \\
        Student & 98.00 & 88.64 & 42.9 \\
        \hline
    \end{tabular}
\end{table}

For the second set of experiments, we train the teacher SNN and use it to perform KD for the student SNN with the proposed loss functions and training methodology. In Table \ref{tab:lossAbalation}, we show the results of using various loss functions and how it affects the classification accuracy of the learned student network. We can observe from this table that the student network achieves a classification performance that is very close to the teacher network after KD using the proposed loss function. Note that the student network achieved this performance without using the ground truth labels but by just being trained using the supervision of the teacher model's spike activation tensors.

\begin{table}[!t]
\caption{Performance comparison of Student SNNs with knowledge distilled from the Teacher model using individual components of the proposed loss function.}
\label{tab:lossAbalation}
\centering
	\begin{tabular}{c|c c c}
        \hline \textbf{Dataset} & \textbf{MNIST} & \textbf{F-MNIST} & \textbf{CIFAR10} \\ \hline
        Teacher & 98.35 & 89.72 & 45.43 \\ \hline
        Full L1 ($L_{L1}$) & 96.20 & 86.99 & 37.90 \\
        Full L2 ($L_{L2}$) & 96.80 & 87.50 & 38.70 \\
	    Full KL ($L_{KL}$) & 97.36 & 88.15 & 39.21 \\
        Sliding L1 ($L_{sL1}$) & 96.09 & 87.28 & 38.31 \\
        Sliding L2 ($L_{sL2}$) & 96.29 & 87.08 & 38.89 \\
        Proposed & \textbf{97.46} & \textbf{88.30} & \textbf{41.28} \\
	\hline
\end{tabular}
\end{table}

\begin{figure}[!t]
\begin{center}
	\begin{subfigure}{0.48\textwidth}
	\includegraphics[width=1\columnwidth]{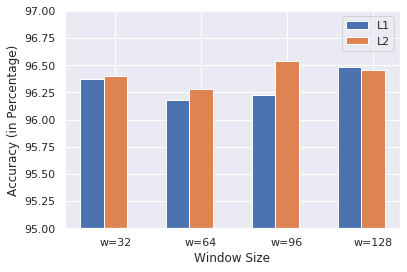}
	\caption{MNIST}
	\end{subfigure}
	\begin{subfigure}{0.48\textwidth}
	\includegraphics[width=1\columnwidth]{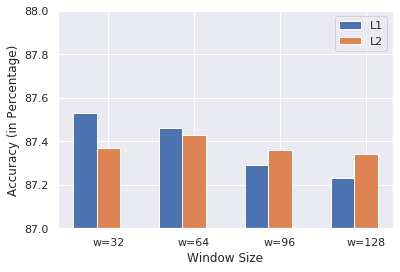}
	\caption{Fashion MNIST}
	\end{subfigure}
	\begin{subfigure}{0.48\textwidth}
	\includegraphics[width=1\columnwidth]{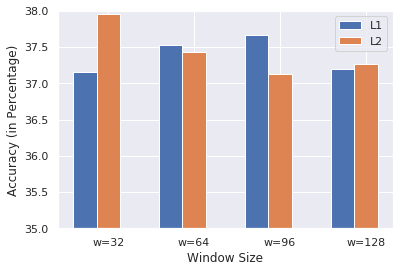}
	\caption{CIFAR10}
	\end{subfigure}
\end{center}
\caption{Effect on classification performance of the student network on varying the sliding window parameter in the loss function for each of the datasets, MNIST, Fashion-MNIST, and CIFAR10.}
\vspace{-5mm}
\label{fig:slidingWindowAbalation}
\end{figure}

\begin{table}[!t]
\caption{Classification performance when using an intermediate TA network for KD from teacher to student.}
\label{tab:taPerformance}
\centering
    \begin{tabular}{c|c c c}
    \hline \textbf{Dataset} & \textbf{MNIST} & \textbf{F-MNIST} & \textbf{CIFAR10}\\ \hline
        Teacher & 98.35 & 89.72 & 45.43 \\ \hline
        T $\rightarrow$ TA & 98.36 & 89.82 & 45.33 \\
        T $\rightarrow$ S & 97.46 & 88.30 & 41.28 \\
        T $\rightarrow$ TA $\rightarrow$ S & \textbf{97.56} & \textbf{88.74} & \textbf{42.38} \\ \hline
\end{tabular}
\end{table}

\begin{figure}[t]
\begin{center}
	\begin{subfigure}{0.48\textwidth}
	\includegraphics[width=1\columnwidth]{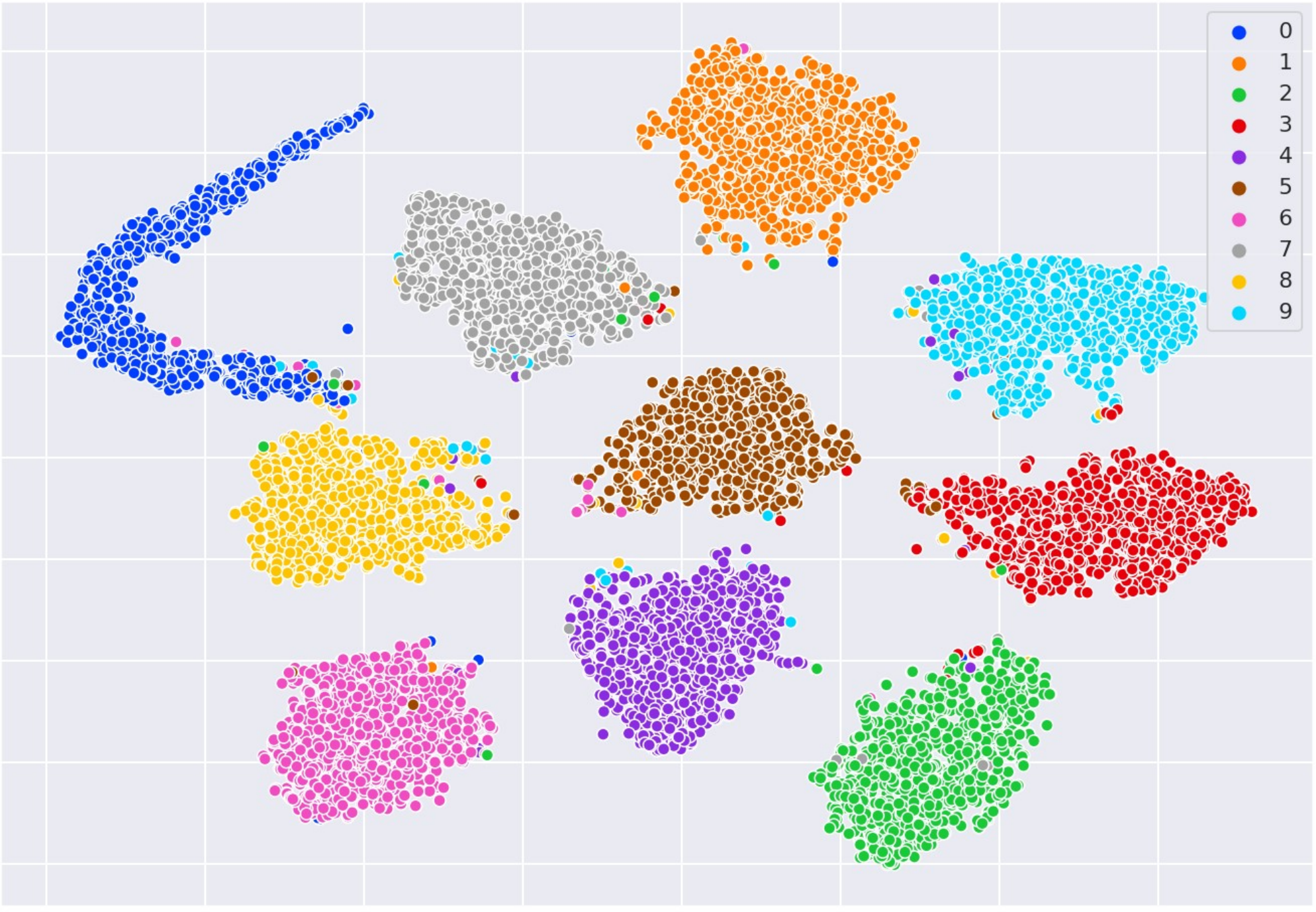}
	\caption{T $\rightarrow$ S}
	\end{subfigure}
	\begin{subfigure}{0.48\textwidth}
	\includegraphics[width=1\columnwidth]{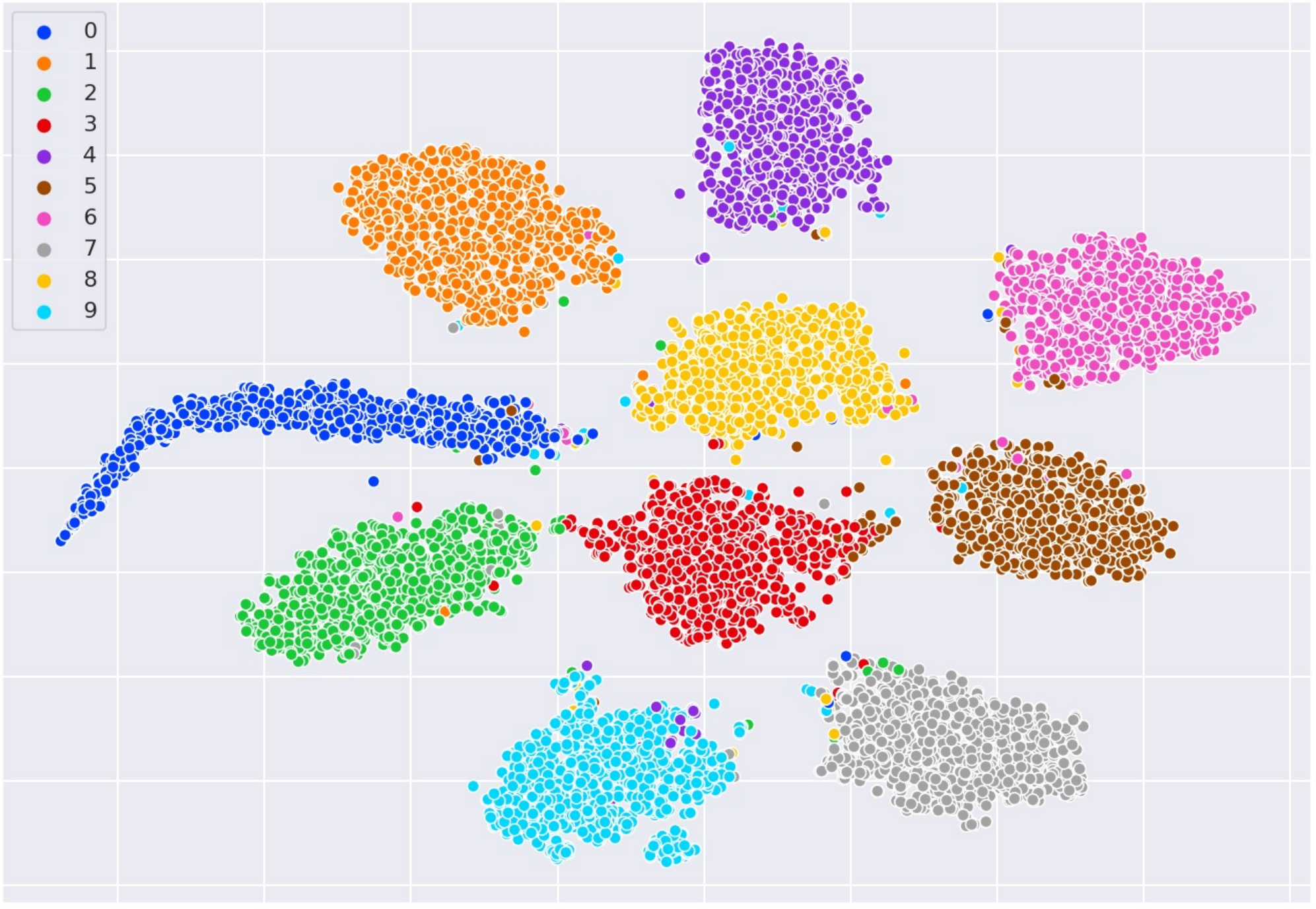}
	\caption{T $\rightarrow$ TA $\rightarrow$ S}
	\end{subfigure}
\end{center}
\caption{TSNE plots for MNIST for knowledge distillation with and without the intermediate TA network. It can be observed that the clusters are better separated when the intermediate training network is used for knowledge distillation.}
\label{fig:mnistTSNEs}
\end{figure}

For the third experiment, we vary the size of the sliding window parameter used i.e., $\Delta$ in Equation \ref{eq:slidingLossCalulation}, for individual components in the proposed loss function calculations. The results of this are shown in Figure \ref{fig:slidingWindowAbalation}. We can observe that varying the sliding window size changes the effectiveness of the loss function and hence, in turn, improves the knowledge distillation performance. It can be observed from the figures that, for sliding L1 loss, we obtain the best results at $\Delta =128$ for MNIST, $\Delta = 32$ for Fashion-MNIST and $\Delta = 96$ for CIFAR10. And  similarly, for sliding L2 loss, we obtain best results at $\Delta = 96$ for MNIST, $\Delta = 64$ for Fashion-MNIST and $\Delta = 32$ for CIFAR10. We observe that for each of the datasets, a different value of $\Delta$ provides the best results, and we accordingly use these best values throughout our experiments.

Finally, we present the fourth set of experiments by including an intermediate network to study the proposed multi-step distillation strategy. We shall call this intermediate network, the Teacher Assistant (TA) network. This multi-stage distillation process involves three steps. First, training a teacher from scratch as was done in the second set of experiments. Next, the Teacher network is used to distill knowledge to the TA network using the proposed methodology and loss function. Finally, this distilled TA model is used to distill knowledge to the Student network using the same methodology and loss function. The results of this experiment are presented in Table \ref{tab:taPerformance}. We can observe that when using this intermediate TA model, the student network achieves a better performance than when it was trained directly with a larger teacher model.

\begin{figure}[t]
\begin{center}
	\begin{subfigure}{0.48\textwidth}
	\includegraphics[width=1\columnwidth]{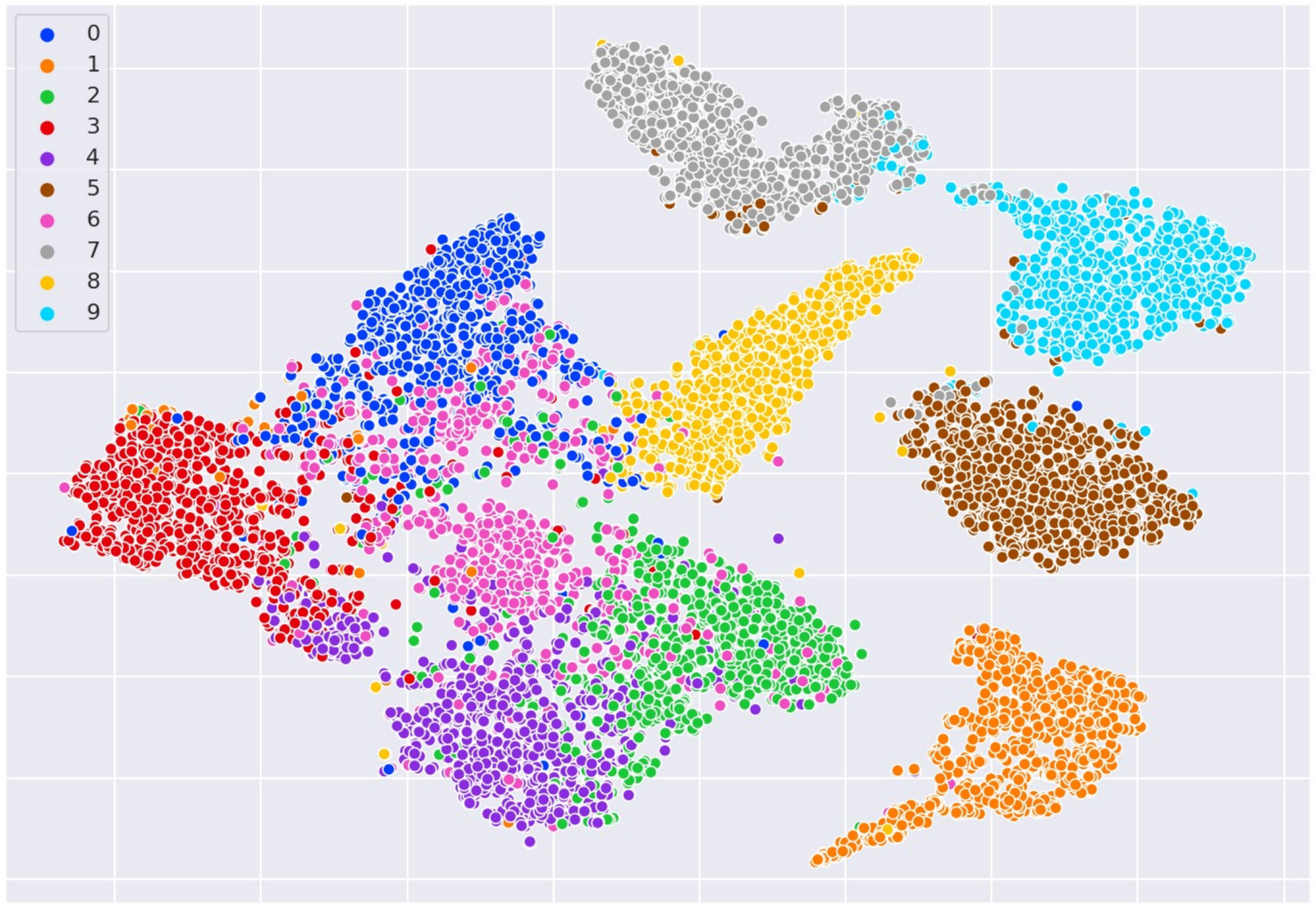}
	\caption{T $\rightarrow$ S}
	\end{subfigure}
	\begin{subfigure}{0.48\textwidth}
	\includegraphics[width=1\columnwidth]{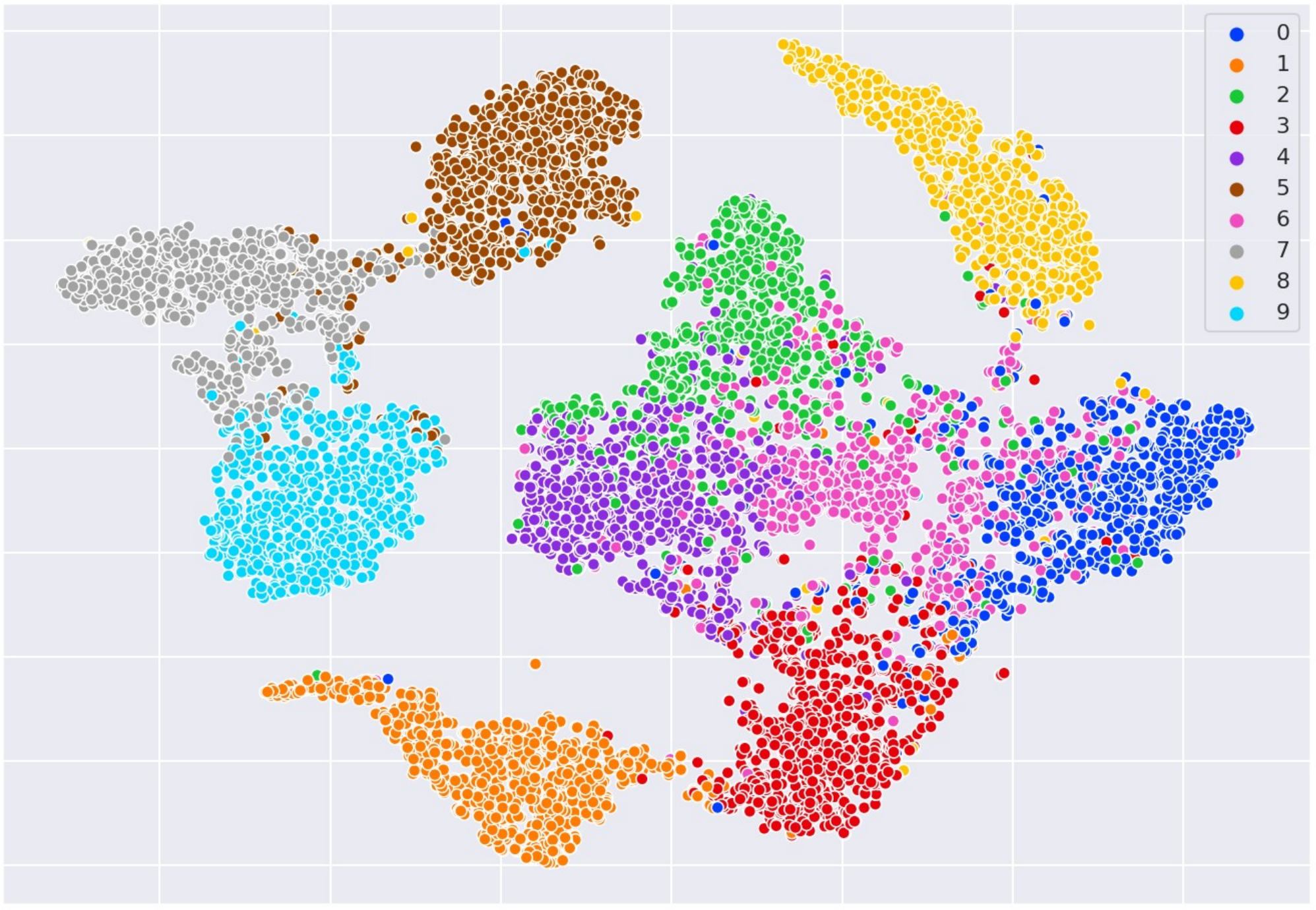}
	\caption{T $\rightarrow$ TA $\rightarrow$ S}
	\end{subfigure}
\end{center}
\caption{TSNE plots for Fashion-MNIST for knowledge distillation with and without the intermediate TA network. It can be observed that the clusters are better separated when the intermediate training network is used for knowledge distillation.}
\label{fig:fmnistTSNEs}
\end{figure}

Finally, for better understanding of the feature separation among classes in each of the student, T $\rightarrow$ S and T $\rightarrow$ TA $\rightarrow$ S SNN models, we also present TSNE plots of the latent features from MNIST and Fashion-MNIST datasets for these respective models in Figures \ref{fig:mnistTSNEs} and \ref{fig:fmnistTSNEs}. It can be observed from these plots that the separation in the classes, as learned by the T $\rightarrow$ TA $\rightarrow$ S is better than the ones learned by the other two models.

\section{Conclusion}
SNNs are energy-efficient neural architectures that benefit from deeper architectures like ANNs. However, deeper models are large and not amenable to deployment and model compression techniques must be developed to enable to make them practical. In this work, we propose a Knowledge distillation that allows transferring the knowledge of the large SNN to smaller one in a disciplined fashion with minimal loss in performance. We provide novel training methodology and objective functions for successfully performing the proposed spike distillation. Along with this, we also propose a multi-step distillation strategy that offers further improvement in performance by using an intermediate TA network. Backed by thorough experiments, we show that using the proposed distillation techniques and objective functions allow an effective spike distillation in SNNs and using a TA network indeed helps in the knowledge distillation process. We expect that the proposed spike distillation procedures can enable practical realization of large SNN models and provide high performance of deeper models while adhering to physical constraints of the available neuromorphic hardware.


\bibliographystyle{IEEEbib}
\bibliography{references}

\end{document}